\documentclass[letterpaper, 10 pt, conference]{ieeeconf}  
\IEEEoverridecommandlockouts                              

\pdfoutput=1
\usepackage{color}
\usepackage{graphicx}
\usepackage{amsmath}
\usepackage{graphicx}
\usepackage{amssymb}
\usepackage{booktabs}
\usepackage{tabularx}
\usepackage{multirow} 
\usepackage{threeparttable}
\usepackage{makecell}
\usepackage{amsmath}
\usepackage{graphicx}
\usepackage{amssymb}
\usepackage{color}
\usepackage{graphicx}
\usepackage{amsmath}
\usepackage{graphicx}
\usepackage{amssymb}
\usepackage{booktabs}
\usepackage{tabularx}
\usepackage{colortbl}
\usepackage{multirow} 
\usepackage{xcolor}
\usepackage{nicematrix}
\usepackage{threeparttable}
\usepackage{makecell}
\usepackage[colorlinks,linkcolor=red]{hyperref}

\usepackage{graphics} 
\usepackage{epsfig} 
\usepackage{times} 
\usepackage{amsmath} 
\usepackage{amssymb}  
\usepackage[linesnumbered,ruled,vlined]{algorithm2e}
\title{\LARGE \bf
Hierarchical End-to-End Autonomous Driving: Integrating BEV Perception with Deep Reinforcement Learning
}

\author{Siyi Lu$^{1,2,3}$, Lei He$^{1,2*}$, Shengbo Eben Li$^{1,2}$, Yugong Luo$^{1,2}$, Jianqiang Wang$^{1,2}$, Keqiang Li$^{1,2}$ 
\thanks{This study is supported by National Natural Science Foundation of China, Science Fund for Creative Research Groups (Grant No.52221005).}
\thanks{$^{1}$Siyi Lu, Lei He, Shengbo Eben Li, Yugong Luo, Jianqiang Wang, Keqiang Li are with School of Vehicle and Mobility, Tsinghua University, Beijing, 100084, China.
        {\tt\small lishbo@tsinghua.edu.cn,
        lyg@mail.tsinghua.edu.cn,
        wjqlws@tsinghua.edu.cn,
        helei2023@tsinghua.edu.cn,
        likq@tsinghua.edu.cn}}
\thanks{$^{2}$Siyi Lu, Lei He, Shengbo Eben Li, Yugong Luo, Jianqiang Wang, Keqiang Li are with State Key Laboratory of Intelligent Green Vehicle and Mobility, Tsinghua University, Beijing, 100084, China.}
\thanks{$^{3}$Siyi Lu is with School of Computer Science and Engineering, Central South University, Changsha, 410083, China. 
        {\tt\small llllsy@csu.edu.cn}}
\thanks{$^{*}$Lei He is the corresponding author.
}
}

\begin{document}

\maketitle
\thispagestyle{empty}
\pagestyle{empty}

\begin{abstract}
End-to-end autonomous driving offers a streamlined alternative to the traditional modular pipeline, integrating perception, prediction, and planning within a single framework. While Deep Reinforcement Learning (DRL) has recently gained traction in this domain, existing approaches often overlook the critical connection between feature extraction of DRL and perception. In this paper, we bridge this gap by mapping the DRL feature extraction network directly to the perception phase, enabling clearer interpretation through semantic segmentation. By leveraging Bird’s-Eye-View (BEV) representations, we propose a novel DRL-based end-to-end driving framework that utilizes multi-sensor inputs to construct a unified three-dimensional understanding of the environment. This BEV-based system extracts and translates critical environmental features into high-level abstract states for DRL, facilitating more informed control. Extensive experimental evaluations demonstrate that our approach not only enhances interpretability but also significantly outperforms state-of-the-art methods in autonomous driving control tasks, reducing the collision rate by 20\%. 
\end{abstract}

\section{Introduction}

End-to-end autonomous driving has garnered significant attention for its ability to unify perception, prediction, and planning into a single, integrated model, offering an alternative to the traditional modular approach \cite{e2ead}. In contrast to the classic pipeline, where separate modules for perception, prediction, and planning are prone to error propagation and high computational complexity \cite{muhammad2020deep,omeiza2021explanations}. Since it can significantly reduce the writing of manual rule codes, end-to-end has gradually become the mainstream trend in the intelligent development of smart connected vehicles \cite{9592698}.

Recent advancements have seen Deep Reinforcement Learning (DRL) applied to end-to-end autonomous driving \cite{zhang2022rethinking}, where the system encodes environmental and vehicular state information into high-dimensional latent feature representations. From these representations, a DRL agent outputs driving strategies for autonomous navigation. However, existing research has often treated feature extraction as an isolated component, without explicitly connecting it to the perception module, which is crucial in traditional driving systems. In this paper, we bridge this gap by mapping DRL’s feature extraction network to the perception phase, and more importantly, utilizing semantic segmentation decoding to interpret the extracted features in a structured manner.

\begin{figure}
    \centering
    \includegraphics[width=0.9\columnwidth]{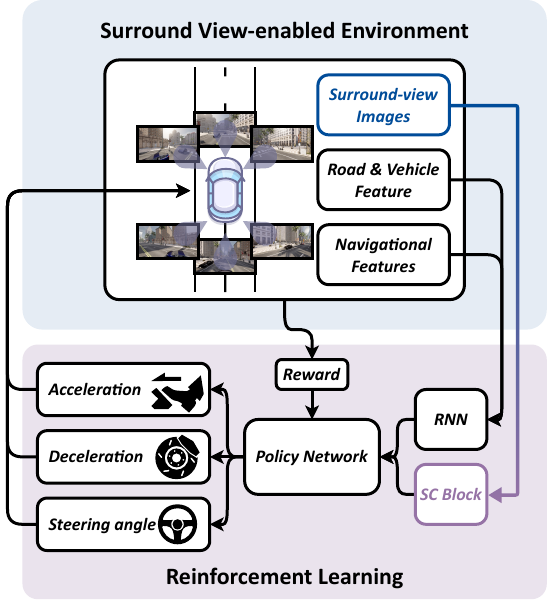}
    \caption{Our perception-driven end-to-end autonomous driving model build on deep reinforcement learning, proposed a feature extraction network based on the bird's-eye view space to process the input surround camera images, output high-dimensional features to the reinforcement learning strategy network, and directly output control information for controlling the vehicle's throttle, brake, and steering wheel.}
    \label{fig:iros_intro}
\end{figure}

Bird’s-Eye-View (BEV) representations have emerged as an effective means for capturing driving scenarios, particularly in urban settings \cite{liang2022bevfusion,liu2023bevfusion,li2023delving,li2022bevformer}. BEV consolidates multi-sensor inputs in a unified three-dimensional space, providing a comprehensive understanding of the vehicle’s surroundings. In our framework, BEV features serve as the input to the DRL policy, which outputs driving control signals. While BEV offers a robust means of feature representation, the process of extracting and translating these features into abstract states suitable for DRL remains challenging. To overcome this, we propose an expressive yet efficient neural network to extract relevant features from BEV inputs and map them directly to the perception phase of the autonomous driving pipeline. By incorporating semantic segmentation into the feature decoding process, we aim to provide a clearer interpretation of the environment, making the DRL agent’s decisions more transparent and informed.

In this paper, we propose a DRL-based end-to-end autonomous driving framework that integrates BEV. The system incorporates input from cameras facing different directions, and construct a BEV representation of the driving environment. A neural network module is designed to extract salient features from the BEV data, capturing relevant information about the surroundings and the vehicle's own state. The extracted BEV features are then fed into a DRL agent, which learns to decode an appropriate driving strategy directly from the sensory inputs, without the need for explicit modeling of the environment. By incorporating the BEV representation, the proposed framework aims to provide the DRL agent with a more holistic and structured understanding of the driving scenario and enhance the agent's ability to reason about the environment and make more informed decisions, leading to improved autonomous driving performance. To the best of our knowledge, this paper is the first solution to combine BEV and deep reinforcement learning for end-to-end autonomous driving.

The main contributions of this paper are as follows: 
\begin{enumerate}
    \item We proposed a feature extraction network in DRL based on bird's-eye view and surround camera inputs to obtain complete environmental information around the vehicle, unify the coordinate system transformation of vehicles, roads, and image inputs, and greatly improve the performance of end-to-end autonomous driving control methods.
    \item Based on semantic segmentation-a classic autonomous driving perception task-we decode the high-dimensional environmental features extracted by the proposed feature extraction network from the surround camera, and visualize the decoded information as other vehicles in the environment, improving the interpretability of DRL.
    \item We evaluate the proposed algorithm on 7 maps in the CARLA, and compare it with the DRL algorithm based on the traditional feature extraction network. The experiment shows that the BEV-based feature extraction network enables the DRL policy network to obtain more rewards, which greatly improves the performance of the autonomous driving control strategy.
\end{enumerate}

\section{Related Work}

\subsection{Traditional Modular Approach}

The traditional modular approach to autonomous driving consists of four main modules: perception, prediction, planning, and control \cite{grigorescu2020survey,10379459}. Each of these modules impacts overall performance.
The early perception module relied on traditional computer vision algorithms, such as edge detection \cite{ziou1998edge}, corner detection \cite{bolte2019towards}, and target tracking \cite{liu2021robust}.
Vehicle trajectory prediction methods based on traditional machine learning include the Kalman filter \cite{lefkopoulos2020interaction, abbas2020adaptive}, Bayesian networks \cite{rio2020vehicle}, and Markov methods \cite{ye2016vehicle}. In contrast, deep learning approaches often use long short-term memory (LSTM) encoder-decoder structures \cite{lin2021vehicle, mo2020interaction}.
The planning module is divided into global and local path planning, calculating trajectory points for the vehicle's low-level controller \cite{claussmann2019review, paden2016survey}. The control module generates safe and reliable real-time instructions based on the driving trajectory \cite{kong2015kinematic}.
A key advantage of this modular design is its interpretability; it breaks down a complex system into independent yet interrelated modules, each focusing on a specific task, making understanding and analysis easier.
 
\subsection{Deep Reinforcement Learning for Autonomous Driving}
Deep Reinforcement learning is a powerful and effective method to obtain end-to-end autonomous driving policies with superior performance. There are many works on end-to-end autonomous driving using reinforcement learning \cite{aradi2020survey, lu2023crowdnav}. Ref.\cite{chen2019model} proposed a framework designed to facilitate model-free deep reinforcement learning within complex urban autonomous driving environments. Ref. \cite{9196730} proposed reinforcement learning within a simulation environment to develop a driving system capable of controlling a full-scale real-world vehicle. The driving policy utilizes RGB images captured from a single camera along with their semantic segmentation as input data. Ref. \cite{8911507} proposed a comprehensive framework for decision-making, amalgamating the principles of planning and learning. This fusion leverages Monte Carlo tree search and deep reinforcement learning to address challenges such as environmental diversity, sensor information uncertainty, and intricate interactions with other road users.

\subsection{Explainability of Autonomous Driving}
Autonomous driving is a high-stakes, safety-critical application. Explainability, which combines interpretability (human comprehensibility) and completeness (exhaustive explanations) \cite{zablocki2022explainability}, is essential for users and traffic participants to trust and accept autonomous systems \cite{zhang2020expectations, du2019look}. Researchers also rely on explainability to optimize and enhance the performance of driving algorithms \cite{bansal2018chauffeurnet, pei2017deepxplore}. As end-to-end autonomous driving develops, the need for interpretability becomes increasingly important. Deep reinforcement learning models, consisting of multiple layers and complex neural networks, often make their decision-making processes and feature representations difficult to understand \cite{chen2024end}. Visual analysis is a key method for enhancing the interpretability of these models \cite{wang2021visual}. This paper proposes a deep reinforcement learning feature extraction network from a bird's eye view (BEV) that integrates perception tasks with feature decoding and visualization, improving both the performance and interpretability of end-to-end autonomous driving algorithms.

\section{Approach}
The end-to-end algorithm framework has attracted great attention in the field of autonomous driving due to its more concise algorithm process and stronger generalization performance. Building on the previous end-to-end autonomous driving approach \cite{e2e-cla}, \cite{chen2021interpretable}, we use multiple cameras mounted on the car as the output of the end-to-end autonomous driving algorithm, and output control signals that control the accelerator, brake, and steering wheel corner. 
\begin{figure*}
    \centering
    \includegraphics[width=0.83\linewidth]{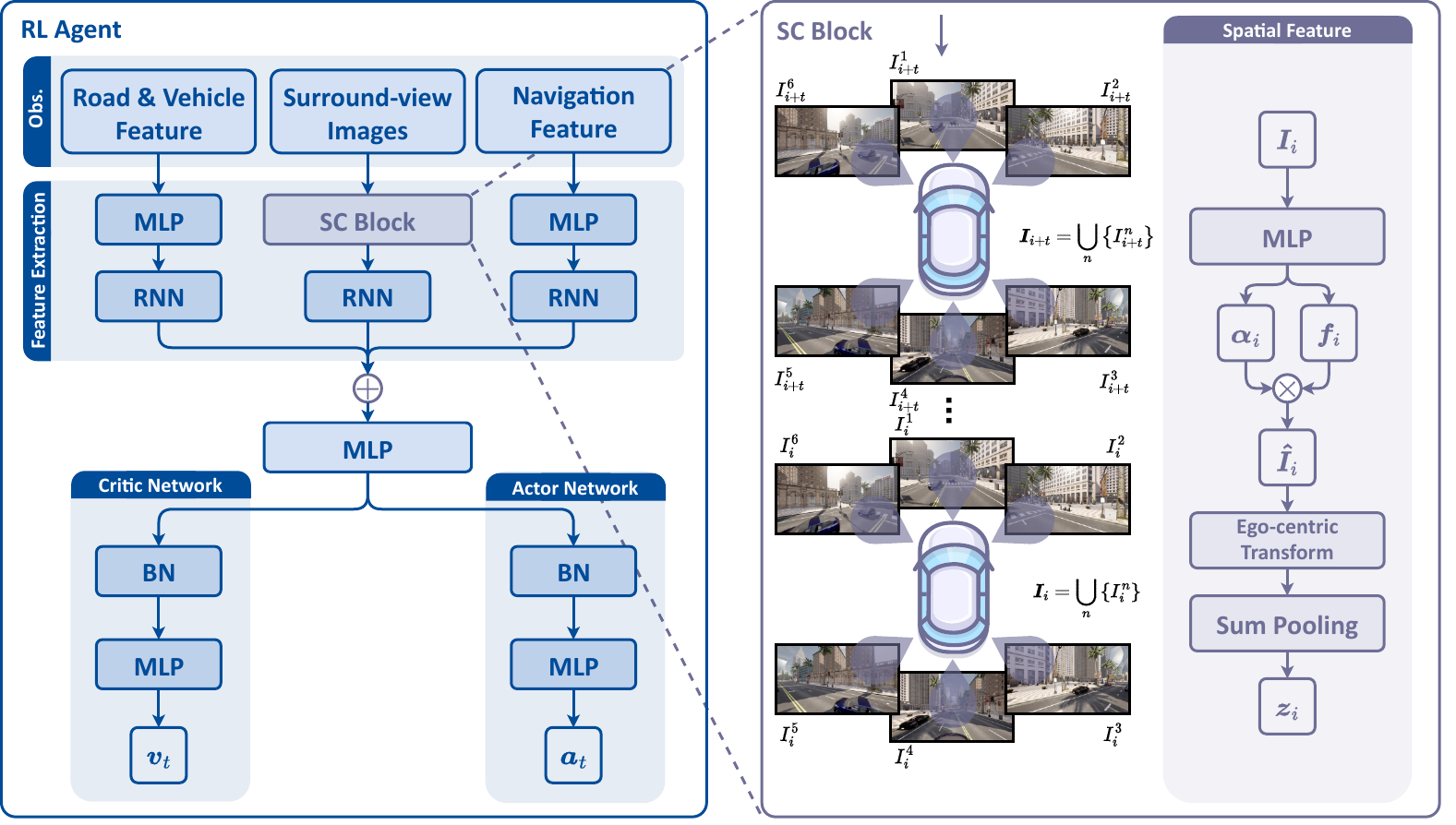}
    \caption{Neural network architecture of the proposed framework. On the left is the architecture of deep reinforcement learning, and on the right is the architecture of the BEV feature extraction network.}
    \label{fig:frame}
\end{figure*}
\subsection{Problem Formulation}
Our work focuses on designing end-to-end autonomous driving algorithms aimed at efficiently reaching a target location while avoiding collisions with other traffic participants. This problem can be modeled as Partially Observable Markov Decision Processes (POMDPs) \cite{9916069}. 
A POMDP can be represented by a tuple $<\mathcal{A},\mathcal{S},\mathcal{R},\mathcal{P},\mathcal{O},\mathcal{Z},\gamma>$, where $\mathcal{A}$ denotes the action space, $\mathcal{S}$ denotes the state space, $R$ represents the reward function, $\mathcal{P}$ is the state transition function, $\mathcal{O}$ is the observation space, $\mathcal{Z}$ is the observation function, and $\gamma$ is the discount factor. In the context of autonomous driving, the state transition function $\mathcal{P}$ and the observation function $\mathcal{Z}$ are often not available in closed-form, rendering this a model-free POMDP problem. The following sections will outline the POMDP formulation for the autonomous driving task.

\begin{itemize}
    \item \textit{State Space} $\mathbf{S}$: we use the CARLA driving simulator~\cite{dosovitskiy2017carla} as the environment for the agent. The state space $\mathbf{S}$ is defined by CARLA and cannot be directly obtained by the agent. At time step $t$, the state of the environment is represented by $s_t$.
    \item \textit{Observation Space} $\mathbf{O}$: similar to \cite{e2e-cla}, the observation of the agent at time step $t$ is $o_t=\bigcup_{m=1}^M{\left\{ \left< I,R,V,N \right> _m \right\}}$, where $I$ is a $6\times3\times128\times352$ image obtained by six RGB camera (front view and rear view, a total of six cameras). $R$ is a 9-dimensional vector representing road features, $V$ is a 4-dimensional vector embedding vehicle features, and $N$ is a 5-dimensional vector containing navigational features.
    \item \textit{Action Space} $\mathbf{A}$: the actions of the agent include acceleration or deceleration (braking) values and steering angle. Their value range are all $\left[ -1,1 \right] $.
    \item \textit{Reward Function} $\mathbf{R}$: the reward function is designed as follows:
    \begin{equation}
        \mathbf{R}\left( s_t,a_t \right) =r_t=\begin{cases}
    	-k_c,\quad\quad\;\;\mathrm{if}\;\mathrm{collision};\\
    	v_m-v_c,\;\;\;\;\;\;\mathrm{if}\;v_c-v_m>0;\\
    	\frac{4v_c\cdot v_s}{\left\| p_c-p_w \right\| _{2}^{2}},\,\,\;\;\mathrm{otherwise},\\
        \end{cases}
    \end{equation}
    where $v_m$, $v_c$, and $v_s$ are respectively the maximum speed limit of the vehicle, the current speed of the vehicle and the similarity of the vehicle with next waypoint $w$. $k_c$ is the penalty value when a collision occurs. $p_c$ and $p_w$ represent the current position of the vehicle and the position of waypoint $w$ respectively.
\end{itemize}


\subsection{Deep Reinforcement Learning-Based Autonomous Driving}
Reinforcement learning has proven to be a powerful technique for solving Partially Observable Markov Decision Processe.By modeling the autonomous driving process as a POMDP, reinforcement learning can be leveraged to derive optimal driving strategies. 
In this paper, we adopt the Proximal Policy Optimization (PPO) \cite{schulman2017proximal} algorithm as the core reinforcement learning method. PPO is known for its stability and efficiency in continuous control tasks, making it suitable for autonomous driving applications. 
The network architecture in our approach adopts the Actor-Critic architecture, and specific details are shown in Fig. ~\ref{fig:frame}. The input to the deep reinforcement learning system includes not only road features (such as road conditions, lane markings, etc.), vehicle features (such as speed, direction, etc.), and navigation feature, but also images from the surround-view camera. Each observation has a separate channel to extract features, and the RNN captures the temporal dependency of the features. Finally, the features of each channel are concentrated and handed over to the Critic network and the Actor network for decision and estimation. 

The feature extractor network of road features and vehicle features is based on the MLP backbone architecture and the feature extractor network of images from the surround-view camera is based on a BEV feature extraction network named SC Block. By integrating the BEV feature extraction network into the actor network, our DRL-based autonomous driving system gains a clearer and more comprehensive understanding of its surroundings, which significantly enhances decision-making performance. The next section will discuss the details of the BEV feature extraction network and its implementation.

\subsection{BEV Feature Extraction Network}
Traditional image feature extraction algorithms usually process in the same coordinate frame as the input image without coordinate transformation. However, other inputs in the perception space of the autonomous driving algorithm are in the BEV space coordinate system. Different coordinate systems will cause errors in the feature fusion process.
The core idea behind this network is to transform raw image data into a 3D representation and project it into the BEV grid similar to \cite{philion2020lift}. The process can be divided into two main steps: Lift and Splat.

\textbf{Lift Step.} In the Lift step, the transformation from a 2D image to a 3D representation is achieved by predicting a depth distribution for each pixel. Given an image $I\in \mathbb{R} ^{3\times H\times W}$, where $H$ and $W$ are the height and width of the image, we learn a depth distribution $\alpha_{h,w}\in\Delta_{|D|-1}$ for each pixel at location $(h,w)$. This distribution is over a predefined set of depth bins $D=\{d_1,d_2,\cdots,d_n\}$. The transformation can be formulated as
\begin{equation}
    I(h,w,d)=\sum_{d\in D}\alpha_{h,w}(d)\cdot f_{h,w}(d),
\end{equation}
where $f_{h,w}(d)$ is the context vector associated with pixel $(h,w)$ at depth $d$, and $\alpha_{h,w}(d)$ is the learned probability that the pixel corresponds to that depth. This step effectively produces a frustum of points for each pixel in the image across multiple depth values, leading to a 3D point cloud. The depth distribution $\alpha_{h,w}(d)$ is learned through a supervised or self-supervised approach, typically using depth ground truth or monocular depth estimation techniques. This allows the network to predict scene geometry, improving the perception of distance and object positioning.

\textbf{Splat Step.} The Splat step projects all the frustums onto a BEV grid. Each frustum is mapped onto a predefined BEV grid using the camera's intrinsic and extrinsic parameters. This allows features from multiple camera views to be fused into a common BEV plane. This step is crucial for generating a consistent representation that facilitates tasks like motion planning and obstacle detection. Sum pooling is employed to assign each 3D point to the nearest voxel in the BEV grid
\begin{equation}
    b_i=\sum_i{EGO_{Ts}\left( u_i \right)}.
\end{equation}
where $u_i$ represents the context vector associated with the 3D point and $EGO_{Ts}$ is the transformation to the ego-vehicle's coordinate system. This step ensures that the 3D points from different views are correctly combined into a consistent BEV representation.

By using this approach, we ensure the model effectively learns to combine data from different camera views into a cohesive BEV representation. The Lift-Splat architecture respects the symmetries in multi-view data, such as permutation invariance and translation equivariance, making it robust to calibration errors and able to fuse inputs from various viewpoints into a unified BEV output.

\subsection{Semantic Segmentation of Latent Feature}
The input feature extraction network in deep reinforcement learning significantly impacts the algorithm's overall performance, but its relationship with the final results is often unclear. To address this, we decode and visualize intermediate features using semantic segmentation to evaluate the performance of our proposed EV Feature Extraction Network.

Semantic segmentation is a key perception task in autonomous driving. It provides essential contextual information, allowing the system to understand road layouts and identify pedestrians, vehicles, and obstacles. In deep reinforcement learning, the input feature extraction network processes information to derive additional environmental features, aligning with the goals of the traditional perception module. We propose a decoding mechanism that utilizes semantic segmentation to transform the latent features from the extraction network into interpretable outputs. In this article, we leverage semantic segmentation to visualize the performance of the BEV Feature Extraction Network.


We use the pre-trained ResNet as the decoder backbone network for semantic segmentation. The latent features output by the BEV Feature Extraction Network are first processed by the convolutional layer for simple feature extraction, and then further processed by the backbone network to obtain higher-level features. Then, through upsampling and feature concatenation, the low-level features are combined to preserve spatial information. This network can effectively generate bird's-eye view semantic segmentation results.

\section{Experimental Results}
In order to verify the performance of representation features in the bird's-eye view space on the autonomous driving method based on reinforcement learning, we tested two of our proposed autonomous driving algorithms based on BEV space-based representation features and reinforcement learning (Ours-3 and Ours-6), using three cameras and six cameras, respectively. It is also compared with other autonomous driving algorithms based on reinforcement learning (DRL and DRL-pan) \cite{e2e-cla}. 

Experiments are conducted with different maps and different traffic participant densities to verify the generalization performance of our proposed algorithm. At the same time, the interpretability of the representation features in the BEV space is visually demonstrated.

\begin{figure}
    \centering
    \includegraphics[width=0.70\linewidth]{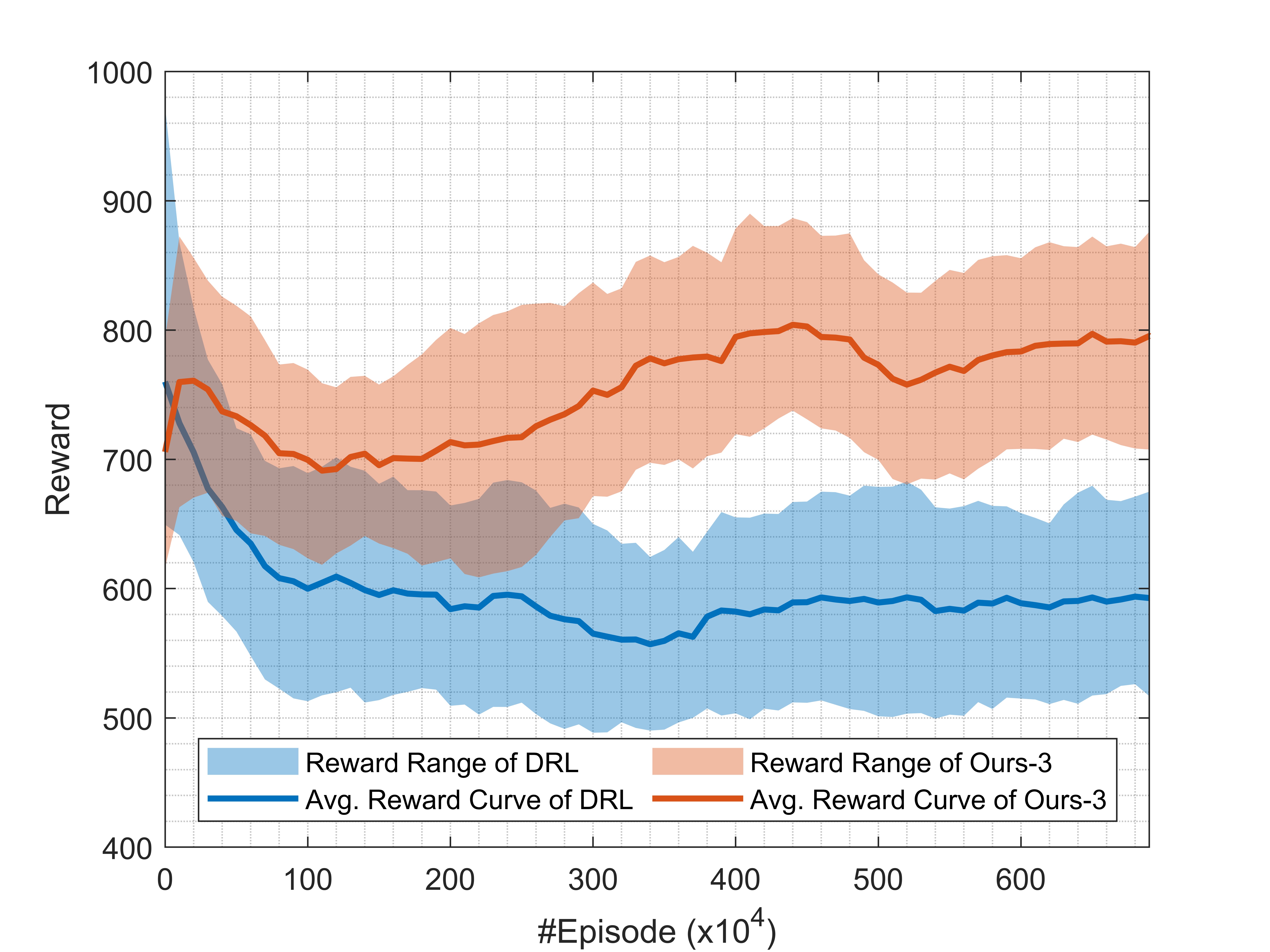}
    \caption{Change curve of the reward function of DRL and Ours-3 method during reinforcement learning training}
    \label{fig:reward}
\end{figure}

\subsection{Experimental Setup}
We use CARLA as a simulator for training and testing autonomous driving algorithms, and autonomous vehicles are equipped with RGB cameras to perceive their surroundings. The DRL method is equipped with three cameras with a 60-degree field of view (FOV), which can observe the image within 180 degrees in front of it. Based on the DRL method, the DRL-pan use three cameras with a FOV of 120 degrees, enabling a 360-degree view of the vehicle's surroundings. The camera setup of the Ours-3 is exactly the same as that of the DRL-pan. Ours-6 uses six cameras with a FOV of 60 to view a 360-degree view of the vehicle's surroundings. Fig. \ref{fig:reward} shows the transformation of the reward function during the training process of the DRL method with three cameras as input and our proposed method.

We selected the Town03 map in CARLA and the traffic flow with low congestion to train four autonomous driving algorithms based on reinforcement learning (DRL, DRL-pan, Ours-3, Ours-6), with 50 pedestrians and 50 cars in the low congestion traffic. The test was conducted in CARLA's Town01~Town07 with a combination of low congestion traffic and high congestion traffic. During autonomous driving, if there is a collision, the mission fails, and conversely, if there is no collision within 128 steps, the mission succeeds.

The evaluation indicators of autonomous driving are \textbf{Collision Rate}, \textbf{Similarity}, \textbf{Timesteps} and \textbf{Waypoint Distance}.\textbf{Collision Rate} refers to the probability of collision while driving, \textbf{Similarity} refers to the average value of the cosine similarity between the direction of vehicle movement and the current vehicle position pointing to the direction of the next planned route waypoint during driving, the \textbf{Timesteps} refers to the driving time before the success or failure of the driving task, and the \textbf{Waypoint Distance} refers to the average value of the distance between the vehicle position and next planned route waypoint during driving. 

\subsection{Evaluation of autonomous driving in different maps}
In order to comprehensively evaluate the performance of our proposed autonomous driving algorithm, we trained the reinforcement learning algorithm on the Town03 map, and verified the performance of the algorithm on seven maps from Town01 to Town07. The results are shown in Table \ref{tab:my-table1}, and our method achieves the best results on most maps and averages for the three indicators Collision Rate, Similarity, and Timesteps. On the average of the 7 maps, ours-6 reduces collision rate by 22\%, improves similarity by 3\%, and increases timesteps by 11.92 compared to the DRL method. On the Waypoint Distance, our method also achieves the best results on 5 maps.This strongly proves that the feature representation in the BEV space enhances the spatial understanding ability of the reinforcement learning agent, which greatly improves the performance of autonomous driving.

Unexpectedly, DRL and $DRL_{pan}$ use the same feature extraction network and the same number of cameras, and the $DRL_{pan}$ method using cameras with a larger field of view can obtain more information in the environment to assist autonomous driving decision-making than DRL. However, the $DRL_{pan}$ method is much worse than the DRL method in three indicators: Collision Rate, Similarity, and Timesteps.The experiment indirectly proves that the expressive power of the feature extraction network shufflenet is limited, thus limiting the performance of the overall reinforcement learning.On the contrary, when our proposed method uses the input of 3 cameras and 6 cameras respectively, the increase in the number of cameras will greatly improve the overall autonomous driving algorithm performance.

\begin{figure*}[t]
    \centering
    \includegraphics[width=1\linewidth]{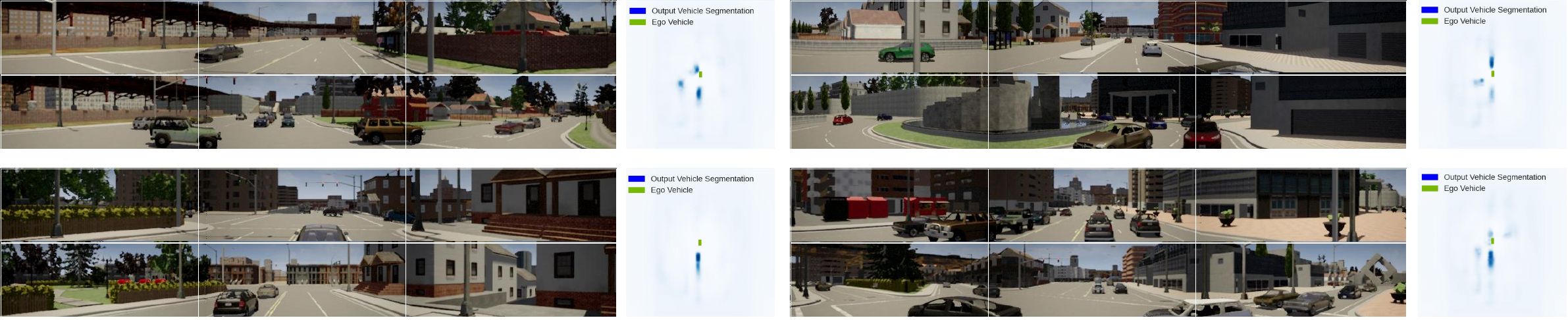}
    \caption{The illustration of the Interpretability of our approach. Each sampling frame is randomly selected from the experiment. The six photos in each sampling frame are taken by a set of surround cameras. The picture on the right is the semantic segmentation result generated by these six pictures.}
    \label{fig:badbev}
\end{figure*}

\begin{table}[]
\centering
\label{tab:my-table1}
\resizebox{\columnwidth}{!}{%
\begin{threeparttable}[b]
\caption{Evaluate the performance of the autonomous driving strategy at low congestion levels in seven scenarios}

\setlength{\tabcolsep}{0.5mm}{
\begin{tabular}{c|c|cccccccc}
\toprule
Matric & Method & Town01 & Town02 & Town03 &  Town04 & Town05 & Town06 & Town07 & Average\\ \midrule
\multirow{4}{*}{Collision Rate $\left( \downarrow \right)$} & DRL &0.37 &0.91 &0.16 &0.26 &0.18 &0.22 &0.31 &0.34  \\
  & DRL-pan &0.44 &0.73 &0.36 &0.52 &0.49 &0.38 &0.67 &0.59  \\
  &\cellcolor{gray!45}Ours-3  &\cellcolor{gray!45}0.50 &\cellcolor{gray!45}0.44 &\cellcolor{gray!45}0.20 &\cellcolor{gray!45}0.21 &\cellcolor{gray!45}0.23 &\cellcolor{gray!45}0.16 &\cellcolor{gray!45}0.35 &\cellcolor{gray!45}0.30   \\
  &\cellcolor{gray!45}Ours-6  &\cellcolor{gray!45}\textbf{\underline{0.14}} &\cellcolor{gray!45}\textbf{\underline{0.12}} &\cellcolor{gray!45}\textbf{\underline{0.11}} &\cellcolor{gray!45}\textbf{\underline{0.11}} &\cellcolor{gray!45}\textbf{\underline{0.11}} &\cellcolor{gray!45}\textbf{\underline{0.11}} &\cellcolor{gray!45}\textbf{\underline{0.16}} &\cellcolor{gray!45}\textbf{\underline{0.12}}   \\ \midrule
  \multirow{4}{*}{Similarity $\left( \uparrow \right)$}  & DRL    & 0.98 &0.00 &0.94 &0.94 &0.96 &0.96 &0.92 &0.81   \\
  & DRL-pan &0.98 &-0.04 &\textbf{\underline{0.98}} &0.16 &-0.04 &0.45 &-0.05 &0.35   \\
  &\cellcolor{gray!45}Ours-3  &\cellcolor{gray!45}0.98 &\cellcolor{gray!45}\textbf{\underline{0.01}} &\cellcolor{gray!45}0.91 &\cellcolor{gray!45}0.21 &\cellcolor{gray!45}0.95 &\cellcolor{gray!45}0.98 &\cellcolor{gray!45}0.95 &\cellcolor{gray!45}0.71   \\
  &\cellcolor{gray!45}Ours-6  &\cellcolor{gray!45}\textbf{\underline{1.00}} &\cellcolor{gray!45}-0.02 &\cellcolor{gray!45}0.95 &\cellcolor{gray!45}\textbf{\underline{0.98}} &\cellcolor{gray!45}\textbf{\underline{0.98}} &\cellcolor{gray!45}\textbf{\underline{1.00}} &\cellcolor{gray!45}\textbf{\underline{0.97}} &\cellcolor{gray!45}\textbf{\underline{0.84}}   \\ \midrule
  \multirow{4}{*}{Timesteps $\left( \uparrow \right)$}  & DRL    & 110.47 &85.71 &123.70 &117.14 &120.86 &119.38 &127.32 &114.94  \\
  & DRL-pan &108.57 &95.66 &114.72 &106.23 &110.43 &111.84 &96.43 &106.27   \\
  &\cellcolor{gray!45}Ours-3  &\cellcolor{gray!45}111.17 &\cellcolor{gray!45}115.14 &\cellcolor{gray!45}121.03 &\cellcolor{gray!45}120.86 &\cellcolor{gray!45}121.67 &\cellcolor{gray!45}123.10 &\cellcolor{gray!45}117.25 &\cellcolor{gray!45}118.60   \\
  &\cellcolor{gray!45}Ours-6  &\cellcolor{gray!45}\textbf{\underline{125.50}} &\cellcolor{gray!45}\underline{\textbf{126.41}} &\cellcolor{gray!45}\underline{\textbf{127.26}} &\cellcolor{gray!45}\underline{\textbf{127.77}} &\cellcolor{gray!45}\underline{\textbf{127.86}} &\cellcolor{gray!45}\underline{\textbf{127.35}} &\cellcolor{gray!45}\underline{\textbf{125.85}} &\cellcolor{gray!45}\underline{\textbf{126.86}}   \\ \midrule
  \multirow{4}{*}{Waypoint Distance $\left( \downarrow \right)$}  & DRL    & 0.96 &29.59 &2.38 &2.50 &2.18 &2.83 &1.47 &5.99   \\
  & DRL-pan &0.99 &\underline{\textbf{19.93}} &\underline{\textbf{1.03}} &2.84 &2.24 &5.54 &1.45 &\underline{\textbf{4.86}}   \\
  &\cellcolor{gray!45}Ours-3 &\cellcolor{gray!45}1.00 &\cellcolor{gray!45}37.53 &\cellcolor{gray!45}2.62 &\cellcolor{gray!45}2.64 &\cellcolor{gray!45}2.16 &\cellcolor{gray!45}3.87 &\cellcolor{gray!45}1.35 &\cellcolor{gray!45}7.31   \\
  &\cellcolor{gray!45}Ours-6 &\cellcolor{gray!45}\underline{\textbf{0.63}} &\cellcolor{gray!45}33.94 &\cellcolor{gray!45}1.76 &\cellcolor{gray!45}\underline{\textbf{1.56}} &\cellcolor{gray!45}\underline{\textbf{1.56}} &\cellcolor{gray!45}\underline{\textbf{1.64}} &\cellcolor{gray!45}\underline{\textbf{0.84}} &\cellcolor{gray!45}5.99   \\  \bottomrule
\end{tabular}}
\end{threeparttable}}
\end{table}

\subsection{Evaluation of autonomous driving in high-congestion environments}
In high-congestion environments, the performance of autonomous driving systems faces greater challenges due to the increased number of dynamic traffic participants. To evaluate the robustness of our proposed BEV-based autonomous driving algorithms under such conditions, we conducted tests using both low and high traffic densities across various CARLA maps. High-congestion scenarios involve 100 pedestrians and 100 vehicles, doubling the complexity compared to the low-congestion settings.

The results of these experiments are shown in Table \ref{tab:my-table1}, where we compare our algorithms (Ours-3 and Ours-6) with the baseline methods (DRL and DRL-pan). As expected, the collision rate increases under high traffic densities for all methods due to the higher likelihood of encountering obstacles and unpredictable behaviors of other traffic participants. However, our proposed methods, particularly Ours-6, demonstrate significantly better collision avoidance capabilities. Ours-6 reduces the collision rate by an average of 18\% compared to DRL across the tested maps, proving that the enhanced spatial understanding from the BEV feature representation is effective even in high-traffic situations.

In addition to collision rate, other evaluation metrics such as similarity, timesteps, and waypoint distance further illustrate the superior performance of our methods in handling congestion. Ours-6 maintains a high degree of similarity, even when surrounded by numerous other vehicles, ensuring the vehicle follows the planned route more precisely. The timesteps achieved by Ours-6 are consistently longer, indicating that the algorithm can successfully navigate through congested environments for extended periods without collisions. Lastly, the waypoint distance remains low for our method, proving that the vehicle stays closer to the optimal route in complex traffic situations.
\begin{table}[]
\centering
\resizebox{\columnwidth}{!}{%
\begin{threeparttable}[b]
\caption{Evaluate the performance of the autonomous driving strategy at high congestion levels in seven scenarios}
\label{tab:my-table1}
\setlength{\tabcolsep}{0.5mm}{
\begin{tabular}{c|c|cccccccc}
\toprule
Matric & Method & Town01 & Town02 & Town03 &  Town04 & Town05 & Town06 & Town07 & Average\\ \midrule
\multirow{4}{*}{Collision rate $\left( \downarrow \right)$} & DRL    & 0.36 &0.88 &0.18 &0.3 &0.22 &0.16 &0.38 &0.35  \\
  & DRL-pan &0.54 &0.71 &0.4 &0.54 &0.51 &0.26 &0.65 &0.52  \\
  & \cellcolor{gray!45}Ours-3  &\cellcolor{gray!45}0.38 &\cellcolor{gray!45}0.39 &\cellcolor{gray!45}0.1 &\cellcolor{gray!45}0.21 &\cellcolor{gray!45}\textbf{\underline{0.08}} &\cellcolor{gray!45}0.23 &\cellcolor{gray!45}0.27 &\cellcolor{gray!45}0.24   \\
  & \cellcolor{gray!45}Ours-6  &\cellcolor{gray!45}\textbf{\underline{0.15}} &\cellcolor{gray!45}\textbf{\underline{0.11}} &\cellcolor{gray!45}\textbf{\underline{0.09}} &\cellcolor{gray!45}\textbf{\underline{0.11}} &\cellcolor{gray!45}\textbf{\underline{0.08}} &\cellcolor{gray!45}\textbf{\underline{0.10}} &\cellcolor{gray!45}\textbf{\underline{0.16}} &\cellcolor{gray!45}\textbf{\underline{0.11}}    \\ \midrule
  \multirow{4}{*}{Similarity $\left( \uparrow \right)$}  & DRL    & 0.98 &0.06 &0.92 &0.96 &\textbf{\underline{0.97}} &0.97 &0.92 &0.82   \\
  & DRL-pan &0.98 &0.01 &\textbf{\underline{0.99}} &0.12 &-0.08 &0.46 &0.05 &0.36   \\
  &\cellcolor{gray!45} Ours-3  &\cellcolor{gray!45}0.98 &\cellcolor{gray!45}-0.02 &\cellcolor{gray!45}0.95 &\cellcolor{gray!45}0.96 &\cellcolor{gray!45}0.96 &\cellcolor{gray!45}0.98 &\cellcolor{gray!45}0.93 &\cellcolor{gray!45}0.82   \\
  &\cellcolor{gray!45} Ours-6  &\cellcolor{gray!45}\textbf{\underline{1.00}} &\cellcolor{gray!45}\textbf{\underline{0.20}} &\cellcolor{gray!45}0.97 &\cellcolor{gray!45}\textbf{\underline{0.99}} &\cellcolor{gray!45}0.96 &\cellcolor{gray!45}\textbf{\underline{1.00}} &\cellcolor{gray!45}\textbf{\underline{0.96}} &\cellcolor{gray!45}\textbf{\underline{0.86}}    \\ \midrule
  \multirow{4}{*}{Timesteps $\left( \uparrow \right)$}  & DRL    & 111.61 &86.44 &122.68 &116.2 &120.36 &122.19 &111.42 &112.98  \\
  & DRL-pan &100.82 &98.29 &108.18 &104.76 &109.85 &120.66 &96.87 &105.63   \\
  &\cellcolor{gray!45} Ours-3  &\cellcolor{gray!45}115.32 &\cellcolor{gray!45}115.95 &\cellcolor{gray!45}124.89 &\cellcolor{gray!45}120.80 &\cellcolor{gray!45}126.06 &\cellcolor{gray!45}120.75 &\cellcolor{gray!45}\textbf{\underline{119.27}} &\cellcolor{gray!45}120.43   \\
  &\cellcolor{gray!45} Ours-6  &\cellcolor{gray!45}\textbf{\underline{121.99 }}&\cellcolor{gray!45}\textbf{\underline{127.13}} &\cellcolor{gray!45}\textbf{\underline{128.22 }}&\cellcolor{gray!45}\textbf{\underline{127.79}} &\cellcolor{gray!45}\textbf{\underline{127.67 }}&\cellcolor{gray!45}\textbf{\underline{127.81}} &\cellcolor{gray!45}118.62 &\cellcolor{gray!45}\textbf{\underline{125.79 }}  \\ \midrule
  \multirow{4}{*}{Waypoint distance $\left( \downarrow \right)$}  & DRL    & 1.00 &33.09 &2.15 &2.13 &2.25 &2.98 &1.43 &6.43   \\
  & DRL-pan &0.98 &\textbf{\underline{18.62}} &\textbf{\underline{0.93}} &297.53 &21.88 &68.20 &12.66 &60.11   \\
  &\cellcolor{gray!45} Ours-3 &\cellcolor{gray!45}1.01 &\cellcolor{gray!45}33.52 &\cellcolor{gray!45}2.16 &\cellcolor{gray!45}2.22 &\cellcolor{gray!45}2.34 &\cellcolor{gray!45}3.70 &\cellcolor{gray!45}1.65 &\cellcolor{gray!45}6.65   \\
  &\cellcolor{gray!45} Ours-6 &\cellcolor{gray!45}\textbf{\underline{0.65}}&\cellcolor{gray!45}32.00 &\cellcolor{gray!45}2.12 &\cellcolor{gray!45}\textbf{\underline{1.63 }}&\cellcolor{gray!45}\textbf{\underline{1.93}}&\cellcolor{gray!45}\textbf{\underline{1.62}} &\cellcolor{gray!45}\textbf{\underline{1.27}} &\cellcolor{gray!45}\textbf{\underline{5.88}}   \\  \bottomrule
\end{tabular}}
\end{threeparttable}}
\end{table}

\subsection{Interpretability}
To assess the interpretability of To implement our proposed framework, we conducted experiments using several randomly selected sampled frames. Fig. \ref{fig:badbev} shows the results of semantically segmented (bird’s eye view) decoding of the latent variables obtained by the BEV feature extraction network. Due to the domain gap between the CARLA simulator and the nuScenes dataset, pre-training on nuScenes alone does not generalize well to the simulated environment in CARLA. The differences in sensor configurations, traffic scenarios, and environment dynamics between these datasets may lead to poor decoding accuracy when transferring to a new domain. Fig. \ref{fig:badbev} shows that the decoding quality is significantly improved after fine-tuning the model using deep reinforcement learning. Fine-tuning adapts the model to the specific features of the CARLA environment, enabling it to generate clearer and more accurate BEV masks. These masks effectively capture the spatial layout of objects and obstacles, providing interpretable insights into the decision-making process.

\section{Conclusion}
In this paper, we present a novel end-to-end control framework for autonomous driving that utilizes a DRL-based approach to integrate perception and control. Our method employs a BEV feature extraction network to convert visual input into latent features, which are then decoded using semantic segmentation for improved interpretability. We tackle the challenges of partial observability by framing the problem as a partially observable
markov decision processe, enhancing the system's ability to make informed control despite incomplete environmental data.
Our approach demonstrates significant advancements in autonomous driving by providing a robust feature extraction and explanation mechanism. It not only improves the interpretability of end-to-end control strategy but also contributes to making autonomous systems more transparent and reliable. Future work will focus on refining depth prediction and camera parameter integration to enhance the accuracy and robustness of BEV feature extraction. Additionally, we plan to explore real-world implementations to evaluate the practical viability of our approach in diverse driving environments.





\bibliographystyle{IEEEtran}
\bibliography{ref.bib}

\end{document}